\documentclass[10pt,twocolumn,letterpaper]{article}

\usepackage{cvpr}
\usepackage{times}
\usepackage{epsfig}
\usepackage{graphicx}
\graphicspath{{./img/}}
\usepackage{amsmath}
\usepackage{amssymb}
\usepackage{ownstyles}
\usepackage{booktabs}
\usepackage{subfig}

\usepackage[pagebackref=true,breaklinks=true,letterpaper=true,colorlinks,bookmarks=false]{hyperref}

 \cvprfinalcopy 

\def\cvprPaperID{****} 

\newcommand{\x}{$\mathbf{x}$~}

\newcommand{\z}{$\mathbf{z}$~}
\newcommand{\zs}{$\mathbf{z^*}$~}

\ifcvprfinal\fi
\begin{document}

\title{VectorDefense: Vectorization as a Defense to Adversarial Examples}

\author{Vishaal Munusamy Kabilan\thanks{This work was done during VMK's internship at Auburn University.}\\
{\tt\small mkvishaal@gmail.com}
\and
Brandon Morris\\
{\tt\small 
blm0026@auburn.edu}
\and
Anh Nguyen\\
{\tt\small anhnguyen@auburn.edu}
\and
\newline
\hspace{3.0em}Auburn University\hspace{3.0em}
}

\maketitle

\begin{abstract}
Training deep neural networks on images represented as grids of pixels has brought to light an interesting phenomenon known as adversarial examples. 
Inspired by how humans reconstruct abstract concepts, we attempt to codify the input bitmap image into a set of compact, interpretable elements to avoid being fooled by the adversarial structures.
We take the first step in this direction by experimenting with image vectorization as an input transformation step to map the adversarial examples back into the natural manifold of MNIST handwritten digits.
We compare our method vs. state-of-the-art input transformations and further discuss the trade-offs between a hand-designed and a learned transformation defense.
\end{abstract}

\section{Introduction}\label{sec:intro}

\begin{figure}
	\hspace{1.5em}
	(a)
	\hspace{2.0em}
	(b)
	\hspace{2.0em}
	(c)
	\hspace{2.0em}
	(d)
	\hspace{2.0em}
	(e)
	\hspace{2.0em}
	(f)
	\newline
	\hspace*{0.5em}
	Clean
	\hspace{0.5em}
	I-FGSM 
	\hspace{1.0em}
	$L_{2}$
	\hspace{1.0em}
	DeepFool
	\hspace{0em}
	JSMA
	\hspace{1.0em}
	$L_{0}$
	\newline
	\hspace*{5.0em}
	\cite{kurakin2016adversarial}
	\hspace{1.5em}
	\cite{carlini2017adversarial}
	\hspace{2.0em}
	\cite{moosavi2016deepfool}
	\hspace{1.5em}
	\cite{papernot2016limitations}
	\hspace{1.5em}
	\cite{carlini2017adversarial}
	\vspace*{-1.2em}
	\begin{center}
		\includegraphics[width=0.48\textwidth]{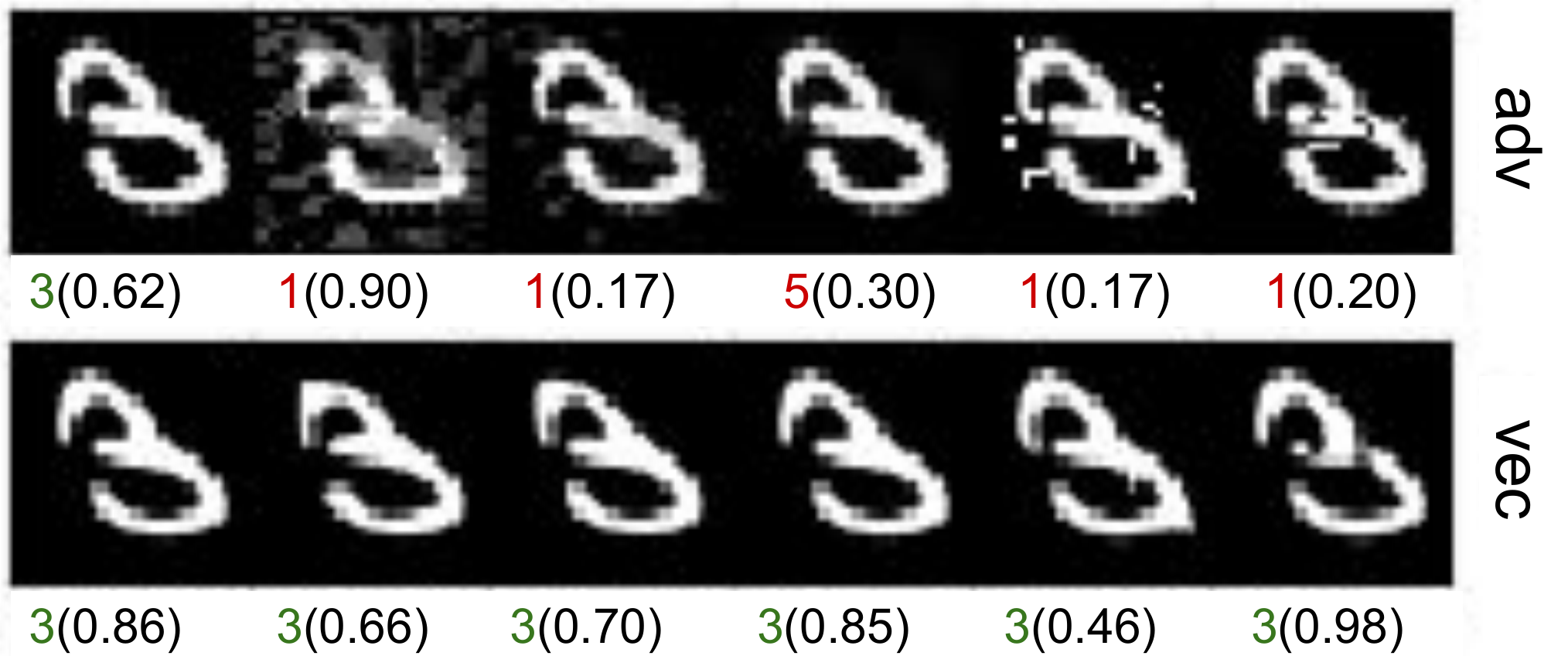}
		\caption{The top row shows the adversarial examples (b-f) crafted for a real test image, here ``3'' (a) via state-of-the-art crafting methods (top labels). The bottom row shows the results of vectorizing the respective images in the row above. Below each image are the predicted label and confidence score from the classifier. 
		All the attacks are targeted to label ``1''
		except for DeepFool (which is an untargeted attack). Vectorization substantially washes out the adversarial artifacts and pulls the images back to the correct labels. See Sec.~\ref{sec:app_ex} for more examples.
		}\label{fig:teaser_image}
	\end{center}
\end{figure}

\begin{figure*}[t]
	\begin{center}
		\includegraphics[width=2.1\columnwidth]{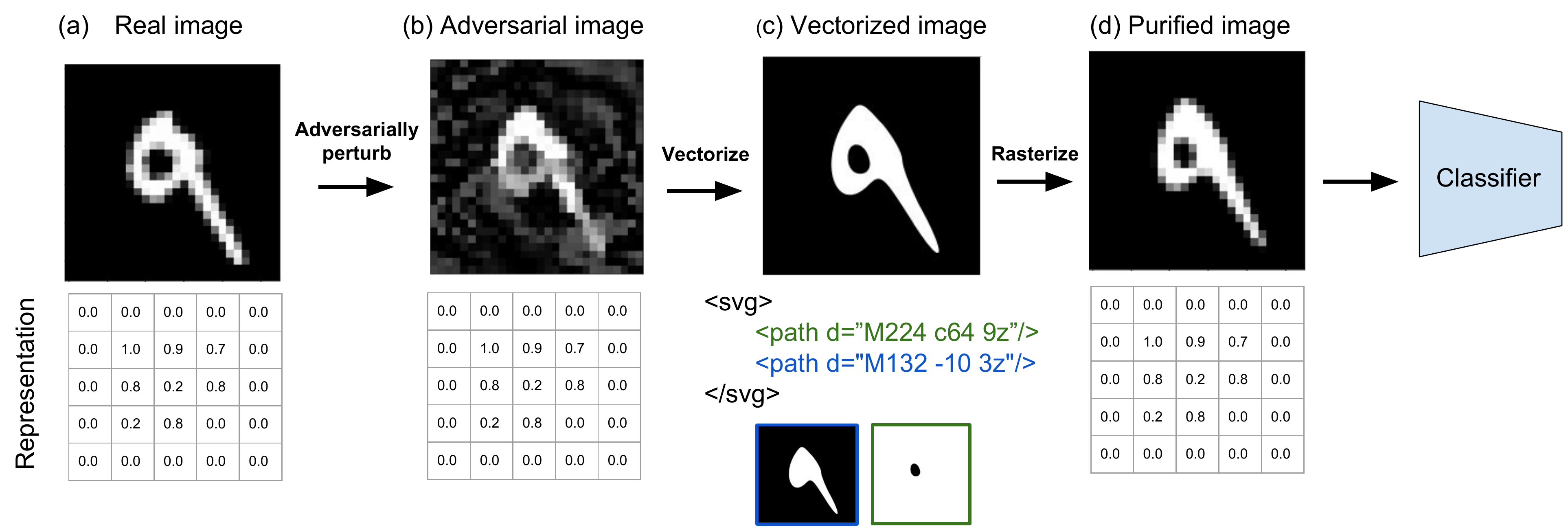}
		\caption{The workflow of VectorDefense.
		Given a real image in the bitmap space (a), an attacker crafts an adversarial example (b).			
		Via vectorization (i.e. image tracing), we transform the input image (b) into a vector graphic (c) in SVG format, which is an XML tree with geometric primitives such as contours and ovals.
		The vector graphic is then rasterized back to bitmap (d) before being fed to a classifier for prediction.
		VectorDefense effectively washes out the adversarial artifacts and pulls the perturbed image (b) back closer to the natural manifold (d).
		}\label{fig:potrace-example}
	\end{center}
\end{figure*}


Deep neural networks (DNNs) have been shown to be extremely vulnerable to
\textit{adversarial examples} (AXs)---the inputs specifically optimized to fool them \cite{szegedy2013intriguing,nguyen2015deep}. An
imperceptible perturbation can be crafted such that when added to the
input of a deep image classifier would change its prediction entirely \cite{szegedy2013intriguing}. These adversarial perturbations can be as minute as one single pixel \cite{su2017one}; computed inexpensively with a single backward pass through the DNN \cite{goodfellow2014explaining} or via black-box optimization algorithms \cite{chen2017zoo,ilyas2017query,papernot2017practical}.



The inability to resist AXs is a general, task- and dataset-agnostic weakness across most machine learning models \cite{akhtar2018threat}.
Importantly, AXs generated to fool one target model \emph{transfers} to fool other unknown models \cite{papernot2017practical}. 
This poses serious implications on the security and reliability of real-world applications, especially the safety-critical domains like autonomous driving.

Recent research suggests that the adversarial problem is an artifact of classifiers being imperfectly trained i.e. having a non-zero error on test data \cite{gilmer2018adversarial,smith2018understanding}. However, perfect generalization is impractical to obtain in real-world settings where the input space is high-dimensional, and our models are trained to minimize the empirical risk over a limited training set. Despite a large body of recent research, the adversarial problem remains largely unsolved \cite{akhtar2018threat,yuan2017adversarial}. A state-of-the-art, effective defense mechanism is \emph{adversarial training} \cite{miyato2015distributional,goodfellow2014explaining}---directly training a model on both clean\footnote{We use \emph{clean} and \emph{real} images interchangeably to refer to real dataset examples (without any perturbations).} and adversarial examples---is not a general solution.
Furthermore, we wish the ideal defense layer to be (1) attack-agnostic, and (2) model-agnostic (e.g. not having to alter or impose additional constraints on an existing image classifier).


In this paper, we explore harnessing \emph{vectorization} (i.e. image tracing) \cite{selinger2003potrace,birdal2014novel}---a simple image transformation method in computer graphics---for defending against AXs. We refer to our method as VectorDefense.
Specifically, we transform each input bitmap image into a vector graphic image (SVG format)---which is composed of simple geometric primitives (e.g. oval and stroke)---via Potrace \cite{selinger2003potrace}, and then rasterize it back into the bitmap form before feeding it to the classifier (Fig.~\ref{fig:potrace-example}).

VectorDefense employs vectorization to smooth out the adversarial perturbations, which are often minute, and local at the pixel level (Fig.~\ref{fig:teaser_image}). 
Here, we attempt to decompose (i.e. vectorize) an input image (e.g. a handwritten digit ``9'') into layers of compact, and interpretable elements (e.g. a stroke and an oval in Fig.~\ref{fig:potrace-example}c) that are resolution-independent. 
We therefore test vectorizing MNIST data, which compose of simple stroke structures, to purify AXs without having to re-train the classifiers.

We make the following contributions:

\begin{enumerate}
	\item We show that VectorDefense is a viable input transformation defense to the problem of AXs. We validate our hypothesis on (1) classifiers trained on MNIST \cite{lecun1998mnist}; (2) across 6 state-of-the-art attack methods; and (3) under both white- and gray-box threat models (Sec. \ref{sec:exp_def}).
	
	\item We compare and show that VectorDefense performs competitively with state-of-the-art \emph{hand-designed} input transformation methods including bit-depth reduction \cite{xu2017feature} and image quilting \cite{guo2017countering} (Sec. \ref{sec:exp_def}). 
	
	
	\item We evaluate the effectiveness of input transformation techniques over the number of pixels allowed to be perturbed (i.e. a budget). That is, we propose budget-aware variants of JSMA \cite{papernot2016limitations} and C\&W $L_0$ \cite{carlini2017adversarial} that craft an AX with a given budget. 
	(Sec. \ref{sec:exp_budget}).

\item We 
compare and contrast VectorDefense with Defense-GAN \cite{samangouei2018defense}, a state-of-the-art input transformation method with a \emph{learned} prior. Interestingly, we found that these two types of approaches can perform distinctively in many cases (Sec. \ref{sec:exp_inter}). 

\end{enumerate}

\begin{figure*}
	\begin{center}
		\includegraphics[width=2.0\columnwidth]{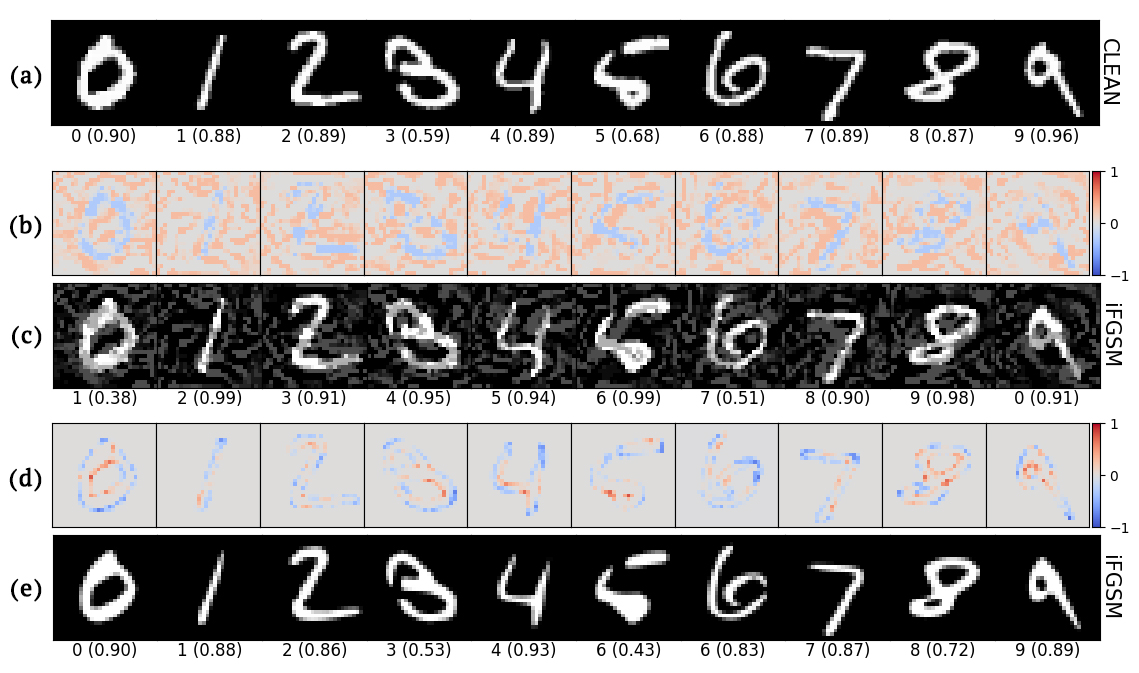}
		\caption{
			Given clean MNIST images (a) which are correctly classified with a label $l$ (here, 0--9 respectively), we add perturbations (b) to produce adversarial examples (c) that are misclassified as $l+1$ (i.e. digit $0$ is misclassified as $1$, etc.) via I-FGSM \cite{kurakin2016adversarial}. 
			VectorDefense effectively purifies the adversarial perturbations in the final results (e), which are correctly classified. Row (d) shows the difference between the original (a) vs. purified examples (e). Below each image we show its predicted label and confidence score. See Sec.~\ref{sec:app_ex} for more examples of VectorDefense vs. other attack algorithms.
		}
		\label{fig:hm_ifgsm}
	\end{center}
\end{figure*}


\section{Adversarial Attacks and Defenses}\label{sec:related-work}

Consider a classifier $f(\cdot)$ that attempts to predict a true label $y$ given
an input image \x. 
An AX is a slight perturbation of an input,
$\mathbf{x}'$, such that $f(\mathbf{x}') \neq y$. 
The AX can be made visually indistinguishable from the original image.
AX crafting algorithms can be also constrained by a norm of the perturbation e.g. $\|\cdot\|_p$ \cite{carlini2017adversarial}.

In this paper, we evaluate our defense under two strong threat models: \textbf{white-box} and \textbf{gray-box} (characterized by the level of knowledge the
adversary has about the victim).
\noindent{White-box:} the attacker has access to the network weights, and also any defense method (here, the input transformation layer). 
\noindent{Gray-box:} the attacker has access to the network parameters, but not the defense method.

%

\subsection{Attacks}\label{sec:attacks}

Many methods for crafting adversarial examples have been proposed \cite{akhtar2018threat}. 
We briefly describe the most relevant gradient-based attacks, which we use to evaluate VectorDefense (Sec.~\ref{sec:experiments}).

A computationally inexpensive procedure is the \textbf{fast gradient
sign method} (FGSM) \cite{goodfellow2014explaining}.
This attack computes the adversarial input through a single backward pass of the network with respect to the input, amounting to a
single-step maximization of the loss function. If the classifier $f(\cdot)$ has
a differentiable loss function $J(\cdot, \cdot)$, calculating the adversarial
input is written as
\begin{equation}\label{eq:fgsm}
  \mathbf{x}' = \mathbf{x} + \epsilon \cdot
  \text{sign}(\nabla_\mathbf{x}J(\mathbf{x}, y))
\end{equation}
where $y$ is the true label and $\epsilon$ determines the strength of the attack
(i.e.\ how much perturbation FGSM is allowed to make). The process can be
adapted to an iterative procedure~\cite{kurakin2016adversarial} that is more
effective with less perturbation:
\begin{equation}\label{eq:ifgsm}
  \mathbf{x}^{(i+1)} = \text{clip}_{\mathbf{x}, \epsilon}(\mathbf{x}^{(i)} +
  \alpha \cdot
  \text{sign}(\nabla_{\mathbf{x}^{(i)}}J(\mathbf{x}^{(i)}, y)))
\end{equation}
where $\mathbf{x}^{(0)} = \mathbf{x}$ is the original image. We refer to this adaptation
as \textbf{iterative fast gradient sign method} (I-FGSM).  
Each iteration, we
take a step of size $\alpha$ in the adversarial direction, always staying
within $\epsilon$ distance of the original input, measured by the $L_\infty$
norm. 
That is, a cap is imposed on the maximum perturbation that can be made at any pixel.
Optimization continues until the image is misclassified or the maximum perturbation has been reached. 


I-FGSM is similar to \textbf{projected gradient descent}
(PGD) attack \cite{madry2017towards} except that PGD starts from a random point within an $\epsilon$-norm ball.

Another gradient-based method is
the \textbf{Jacobian Saliency Map Algorithm}
(JSMA) \cite{papernot2016limitations} which relies on the saliency map computed from the backpropagation.
The saliency map shows how sensitive the network's prediction is to the input pixels. 
In each step, JSMA selects two pixels to
perturb, and repeats the process until a perturbation limit is reached or the DNN prediction becomes incorrect.

The \textbf{DeepFool} algorithm \cite{moosavi2016deepfool} finds an AX by iteratively projecting $\mathbf{x}$ onto a linearized approximation of the decision boundary of classifier $f(\cdot)$.
In each iteration, DeepFool performs the below update:
\begin{equation}\label{eq:deepfool}
  \mathbf{x}^{(i+1)} = \mathbf{x}^{(i)} - \epsilon \cdot
  \frac{f(\mathbf{x}^{(i)})}{\lVert \nabla_{\mathbf{x}^{(i)}}
    f(\mathbf{x}^{(i)}) \rVert_2^2} \nabla_{\mathbf{x}^{(i)}}
    f(\mathbf{x}^{(i)})
\end{equation}
where again $\mathbf{x}^{(0)} = \mathbf{x}$.


\textbf{Carlini and Wagner (C\&W) attack} \cite{carlini2017towards} instead uses Adam optimizer \cite{kingma2014adam} and incorporates a constraint on the adversarial perturbation in three different ways: 
(1) limiting the amount of changes in $L_2$ distance from the original image; (2) limiting the maximum amount of changes to any pixel via $L_\infty$; (3) limiting the number of pixels that can be perturbed (i.e. a budget) via $L_0$.


To study the effectiveness of the input transformation defenses over perturbation budget settings, we modify JSMA and C\&W $L_0$ to take in an additional input parameter that specifies the number of pixels that can be modified. We refer to these algorithms as (1) \textbf{Budget-aware JSMA} and (2) \textbf{Budget-aware C\&W $L_0$ } respectively:

\begin{itemize}
	\item In every step, budget-aware JSMA selects a pair of pixels to perturb until the budget limit is reached regardless of whether the image is misclassified (i.e. $f(\mathbf{x}') \neq y$).
	
	\item In contrast, budget-aware C\&W $L_0$ maintains a set of all pixels that can be perturbed (initially the entire image), and then iteratively shrinks this set until its size equals a given budget.
When it is not possible to find an AX at the given budget, the original image is returned.
\end{itemize}

\subsection{Defenses}\label{subsec:related-work}

A number of defense strategies have been proposed to stymie the effectiveness
of adversarial perturbations \cite{akhtar2018threat}. Here we describe existing 
\emph{input transformation} schemes for later comparing with our VectorDefense.
Input transformation defenses seek to ``purify'' an AX i.e. removing the adversarial perturbations (often small and local) while maintaining the features necessary for correct classifications.
Input transformations can be categorized into those with hand-designed vs. learned priors.
%

Under \textbf{hand-designed} priors, many methods have been proposed \cite{akhtar2018threat}. 
An effective technique is bit-depth reduction \cite{xu2017feature}, which reduces the degrees of perturbation freedom
within each pixel by quantizing the image.
Image quilting \cite{guo2017countering} instead replaces patches of an image with
similar clean patches extracted from a database.


There are input transformation defenses that instead harness a \textbf{learned prior} in the form of Generative Adversarial Networks (GANs) \cite{samangouei2018defense,meng2017magnet,shen2017ape,ilyas2017robust}, and PixelCNN \cite{song2017pixeldefend}.

Among those, we directly compare VectorDefense with a state-of-the-art method called Defense-GAN \cite{samangouei2018defense}, which harnesses a GAN \cite{goodfellow2014generative} to purify AXs. 
Basically, the idea is similar to conditional iterative image generation techniques \cite{nguyen2017plug,nguyen2016synthesizing,zhu2016generative}. 
First, we train a generator $G(\cdot)$ to map a latent code \z from a simple prior distribution (e.g. Gaussian) to an image \x in a target distribution (e.g. handwritten digits in MNIST) which the classifier is also trained on. 
Then, we search for a latent code \zs such that the resulting image matches a target (adversarial) image \x  i.e. $G(\mathbf{z^*}) \approx \mathbf{x}$. The final image found $G(\mathbf{z^*})$ is then fed to the classifier for prediction.
Interestingly, this method is not completely robust. Since $G$ is not guaranteed to be perfectly trained and therefore, one can find AXs directly in the generated distribution \cite{athalye2018obfuscated,ilyas2017robust}. 
Overall, these defenses, and others that do not explicitly use input transformations, are
still imperfect at resisting AXs \cite{carlini2016defensive,carlini2017adversarial,athalye2018obfuscated,akhtar2018threat}.


This work explores a novel, intuitive method that translates an input image into contours and other simple geometric shapes in an attempt to purify AXs. 
We view VectorDefense as a stepping stone towards decomposing images into compact, interpretable elements to solve the adversarial problem.


\section{Methods}\label{sec:method}



VectorDefense is an input transformation defense that aims to remove the adversarial artifacts in an AX, while preserving the
class-specific features needed for classification.
Specifically, we utilize Potrace vectorization algorithm~\cite{selinger2003potrace} to translate an input bitmap image into a vector graphic. Then we rasterize the vector graphic back to a bitmap before classification (Fig.~\ref{fig:potrace-example}).
We first describe Potrace (Sec.~\ref{sec:potrace}) and then how the defense method works (Sec.~\ref{sec:vectordefense}).

\subsection{The Potrace Algorithm}
\label{sec:potrace}


Potrace is an algorithm to convert a black-and-white bitmap image (i.e. a binary image), into an algebraic description of its contours, typically in
B\'ezier curves \cite{selinger2003potrace}. 
Therefore, a vector image is often smooth-edged and does not have pixelation effect (Fig.~\ref{fig:potrace-example}c vs. d).
This description is preferred in graphic design since it is scale-invariant. We chose to experiment with Potrace because the algorithm is simple, efficient and has been shown effective for handwritten images \cite{selinger2003potrace}.

At a high level, the algorithm
 works by tracing polygons onto the bitmap image
based on intensity differences, then optimizing curves to match the contours of
those polygons \cite{selinger2003potrace}. There are four steps in Potrace:

\noindent\textbf{Step 1:} An input bitmap image is decomposed into a set of paths that are formed by drawing boundaries that
separate black and white regions. 

\noindent\textbf{Step 2:} Potrace approximates each path found by a polygon.
The polygons are constructed with the constraints of having the fewest number of edges while matching their respective paths.

\noindent\textbf{Step 3:} Potrace transforms each polygon into a smooth vector outline (i.e. a set of contours).

\noindent\textbf{Step 4:} Potrace joins adjacent B\'ezier curve segments together into a more compact vector file, though this often has imperceptible visual effects on the final output. 

\subsection{Vectorization as a Defense}
\label{sec:vectordefense}


We describe the components of the VectorDefense pipeline that are hypothesized to be effective in purifying AXs.

\noindent\textbf{Binarization} Vectorization is an image tracing algorithm based on a color palette \cite{selinger2003potrace}.
Here, we choose to use only black and white colors for tracing i.e. we
binarize an image first before feeding into Potrace.
The intuition that binarizing images effectively reduces the input space dimension where adversarial perturbations could be made has also been confirmed in a concurrent work of bit-depth reduction \cite{xu2017feature} (here, we reduce images into 1-bit).

\begin{figure}[h]
	\begin{center}
		\includegraphics[width=0.48\textwidth]{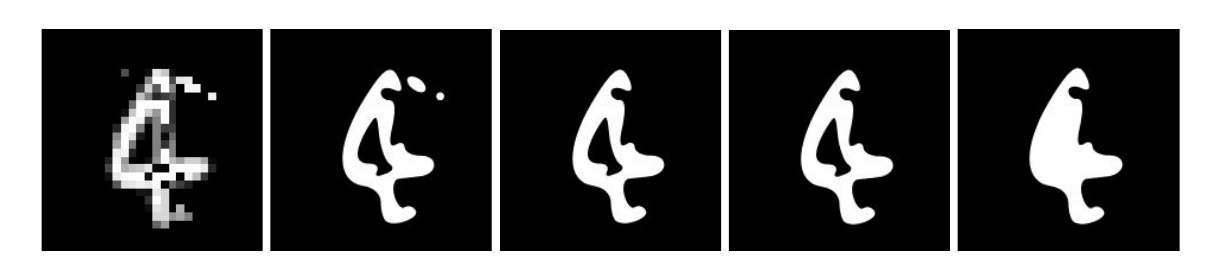}
				\hspace*{0.5em}
		(a) \textbf{AX}
		\hspace{2.0em}
		(b) \textbf{0}
		\hspace{2.0em}
		(c) \textbf{5}
		\hspace{2.0em}
		(d) \textbf{10}
		\hspace{1.0em}
		(e) \textbf{20}
		\hspace{1.0em}
		\caption{
Despeckling removes small color blobs by dropping all the paths that consists of fewer than $t$ pixels. We show the results of Potrace vectorization on an adversarial example crafted by C\&W $L_0$ method \cite{carlini2017adversarial} (a) with different values of $t=0,5,10,20$. $t=0$: no despeckling.
		}
		\label{fig:despeckle_image}
	\end{center}
	
\end{figure}

\noindent\textbf{Despeckling}
An adversarial example often has many speckles i.e. small color blobs that are not part of the digit but injected to fool the classifier (Fig.~\ref{fig:teaser_image}; top row).
Despeckling is an attempt to remove these speckles by dropping all the paths (returned from step 1 in Potrace) that consists of fewer than $t$ pixels.
We empirically find that $t=5$ works the best in removing adversarial speckles  (Fig.~\ref{fig:despeckle_image}).

\noindent\textbf{Smoothing} Real handwritten digits often possess a large amount of smoothness (Fig.~\ref{fig:hm_ifgsm}a), which is corrupted by adversarial perturbations (Fig.~\ref{fig:hm_ifgsm}c).
The steps 1--3 of Potrace vectorize an image into a set of contours, intuitively smoothing out the noisy perturbations (Fig.~\ref{fig:despeckle_image}a~vs.~b). Therefore, vectorization is also effectively a hand-designed smoothness prior e.g. total variation \cite{guo2017countering}.




\section{Experiments \& Results}
\label{sec:experiments}

We evaluate VectorDefense as an input transformation layer to help a state-of-the-art DNN classifier trained on MNIST correctly classify AXs created by \textbf{6 state-of-the-art gradient-based attack methods}: I-FGSM, C\&W $L_2$, PGD, DeepFool, JSMA and C\&W $L_0$.

We compare VectorDefense against two \emph{hand-designed} input transformation methods: image quilting \cite{guo2017countering}, and bit-depth reduction \cite{xu2017feature} because of the following reasons.

Image quilting was shown to be one of the most effective input transformation techniques on natural images, i.e. ImageNet \cite{guo2017countering}. Here, we implement image quilting for MNIST (details in Sec. \ref{sec:experimental_setup}).
Bit-depth reduction obtained state-of-the-art results on MNIST, especially by converting the images into 1-bit (rather than other bit-depth levels) \cite{xu2017feature}.
In addition, comparing VectorDefense (which includes binarization) vs. binarization alone enables us to highlight the effectiveness of the entire vectorization transformation.

In addition, we also compare and contrast with a \emph{learned} input transformation method: Defense-GAN \cite{samangouei2018defense} to shed more light into the pros and cons of hand-designed vs. learned priors.



\subsection{Experiment setup}
\noindent\textbf{Datasets and Networks} The victim model is a DNN classifier from \cite{tramer2017ensemble} trained on the MNIST dataset \cite{lecun1998mnist}. The network architecture is described in Table \ref{table:mnist_arch}. 

We choose MNIST for two reasons: 
(1) research is still needed to understand and defend against AXs even at simple, low-dimensional problems \cite{akhtar2018threat,gilmer2018adversarial}, which can inform defenses in a high-dimensional space (e.g. ImageNet) \cite{gilmer2018adversarial};
(2) the handwritten digits are made of simple strokes,
which can be vectorized well by Potrace into decomposable geometric primitives.

\noindent\textbf{Implementation} 
All of the attack algorithms are from the cleverhans library \cite{papernot2016cleverhans}, except C\&W $L_0$ which is
from the code released by \cite{carlini2017towards}. Defense methods: Image quilting,
Defense-GAN and BPDA are from the code released by \cite{athalye2018obfuscated}.
We implemented the rest of the algorithms, and the code to reproduce our results is at \url{https://github.com/VishaalMK/VectorDefense}.

\noindent\textbf{Hyperparameters} Additional hyperparaters for our experiments are reported in Sec.~\ref{sec:experimental_setup}.

\subsection{VectorDefense in a Gray-Box Setting}
\label{sec:exp_def}

\begin{table*}[t]
	\small
	\centering
	\begin{tabular}{p{11.5em}llllllll}
		\toprule
		& (a) & (b) &  (c) & (d) & (e) & (g) & (h)\\
		& Clean & I-FGSM \cite{kurakin2016adversarial} &  C\&W$L_2$ \cite{carlini2017adversarial} & DeepFool \cite{moosavi2016deepfool} & PGD \cite{madry2017towards} & C\&W$L_0$ \cite{carlini2017adversarial} & JSMA \cite{papernot2016limitations}\\
		\midrule
		No defense     & $99.45$ & $0.40$ & $0.00$ & $0.30$ & $0.10$ & $0.00$ & $0.00$ \\
		\midrule
		Defense-GAN \cite{samangouei2018defense}    & $98.30$ & $96.00$ & $97.50$ & $97.60$ & $97.00$ & $94.00$ & $93.00$ \\
		\midrule
		Bit-depth Reduction \cite{xu2017feature} & $99.30$ & $97.40$ & $96.10$ & $99.30$ & $98.20$ & $31.00$ & $32.00$ \\
		Quilting \cite{guo2017countering}       & $99.30$ & $93.10$ & $96.90$ & $99.00$ & $96.00$ & $47.20$ & $65.00$ \\
		\textbf{VectorDefense} (this)  & $98.60$ & $95.60$ & $91.80$ & $97.40$ & $96.20$ & $60.00$ & $94.00$ \\
		\bottomrule
	\end{tabular}
	\caption{
		\textbf{Input transformation defenses against gray-box attacks.} We
		compare our VectorDefense method against state-of-the-art input transformation defenses on 6 attack
		algorithms (b--h).
	}
	\label{tbl:gray-results}
\end{table*}


We evaluate VectorDefense under the gray-box threat model, which has strong assumptions to defend against i.e. the attacker does have knowledge of the victim model, but not that of the defense mechanism being employed.

\noindent\textbf{Experiment} For each of the 6 attack methods, we compute 1000 AXs on the target DNN from 1000 random test set images. We feed the AXs through four defense methods: (1) Bit-depth reduction; (2) Image quilting; (3) VectorDefense; and (4) Defense-GAN; to produce purified images, which are then fed to the victim DNN for classification.

All hyperparameters for the 6 attacks are default from cleverhans and the crafted AXs cause the victim DNN to obtain at most only 0.4\% accuracy (Table~\ref{tbl:gray-results}; No defense).


\noindent\textbf{Results} We found that
	VectorDefense performed similarly to all three defense methods across four attack methods (I-FGSM; C\&W $L_2$, DeepFool, and PGD).
	Table~\ref{tbl:gray-results} reports the accuracy scores for all 4 defenses vs. 6 attack methods.

\begin{figure}[th]
	\begin{center}
		Number of pixels perturbed by JSMA attack
	\end{center}
	\hspace*{0.5em}
	(a) \textbf{8}
	\hspace{1.0em}
	(b) \textbf{16}
	\hspace{1.0em}
	(c) \textbf{24}
	\hspace{0.8em}
	(d) \textbf{32}
	\hspace{0.5em}
	(e) \textbf{40}
	\hspace{0.5em}
	(g) \textbf{48}
	\vspace*{-1.2em}
	\begin{center}
		\includegraphics[width=0.48\textwidth]{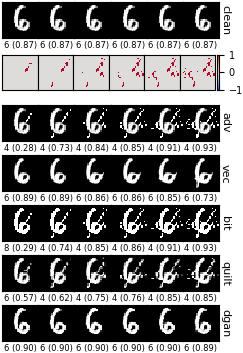}
		\caption{Qualitative comparison of input transformations on budget-aware JSMA.
		Clean images (\emph{clean}) are added to adversarial perturbations (row 2) to produce adversarial images (\emph{adv}).
		Each column (a--g) corresponds to a specific budget setting denoted on top. 
		Rows 4--7 show the results of VectorDefense (\emph{vec}), bit-depth reduction \cite{xu2017feature} (\emph{bit}), image quilting \cite{guo2017countering} (\emph{quilt}) and Defense-GAN \cite{samangouei2018defense} (\emph{dgan}) respectively. 
		VectorDefense effectively maps adversarial images back to the natural manifold.
		}\label{fig:jsma_qual}
	\end{center}
\end{figure}

Specifically, for C\&W $L_0$ and JSMA attacks, VectorDefense substantially outperformed existing hand-designed transformation methods (Table~\ref{tbl:gray-results}h). 
Qualitative comparisons also confirmed our quantitative result (Fig.~\ref{fig:jsma_qual}c).

Compared to other attack methods, C\&W $L_0$ poses a distinct challenge for all input transformation methods by explicitly deleting input features (here, setting many pixels to black in the image; see Fig.~\ref{fig:hm_cwl0}c \& Fig.~\ref{fig:teaser_image} top row).
Under such a strong attack, Defense-GAN substantially outperformed all hand-designed methods (Table~\ref{tbl:gray-results}g).
This result highlighted a big performance difference between a hand-designed vs. a strong learned input transformation method and informed our experiment in the next section.

See Sec.~\ref{sec:app_ex} for more qualitative results of VectorDefense defending against each attack method considered.




\begin{figure}[th]
	\hspace*{1.1em}
	(a)
	\hspace{2.1em}
	(b)
	\hspace{2.2em}
	(c)
	\hspace{2.0em}
	(d)
	\hspace{1.8
		em}
	(e)
	\hspace{1.7em}
	(f)
	\vspace*{-1.2em}
	\begin{center}
		\includegraphics[width=0.48\textwidth]{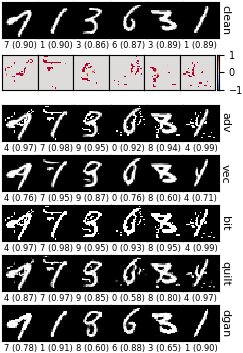}
		\caption{
			We show 6 cases (a--f) to highlight the qualitative difference between how hand-crafted input transformations (VectorDefense (\emph{vec}), bit-depth reduction (\emph{bit}), image quilting (\emph{quilt})) purify adversarial examples (AXs) compared to 
			Defense-GAN (\emph{dgan}). 
			AXs were crafted by perturbing the original images at 40 pixels via Budget-aware JSMA. 
			While VectorDefense transforms AXs into arguably human interpretable 4, 7, 9 or 0 (\emph{vec}; a--d), Defense-GAN interestingly pulls the AXs back into 7, 1, 8, and 6 (bottom row).
		}
		\label{fig:interpretable_qual}
	\end{center}
\end{figure}

\subsection{Gray-Box: Budget-Aware Attack Algorithms}\label{sec:exp_budget}
To study the performance of
input transformations against state-of-the-art $L_0$ attacks, we make use of (1) Budget-aware JSMA and (2) Budget-aware C\&W $L_0$ (described in Sec.~\ref{sec:attacks}). 
These attacks explicitly take in as input a \emph{budget}---the number of pixels the algorithm is allowed to modify.

\noindent\textbf{Experiment}
We sweep across 6 budget settings, from 8 to 48 pixels (in increments of 8 pixels).
The existence of an AX is not guaranteed for every (original image, target label) pair, across every budget setting.
Across all the budget settings, we therefore evaluate the defenses on a subset of 80 successful, targeted AXs, chosen from a set of one thousand AXs generated via each budget-aware algorithm. 
A successful AX is one that is (1) misclassified as a target class; and (2) crafted using the exact given budget. 

\begin{figure*}%
	\centering
	\subfloat[Budget-aware JSMA Attack]{{\includegraphics[width=1.0\columnwidth]{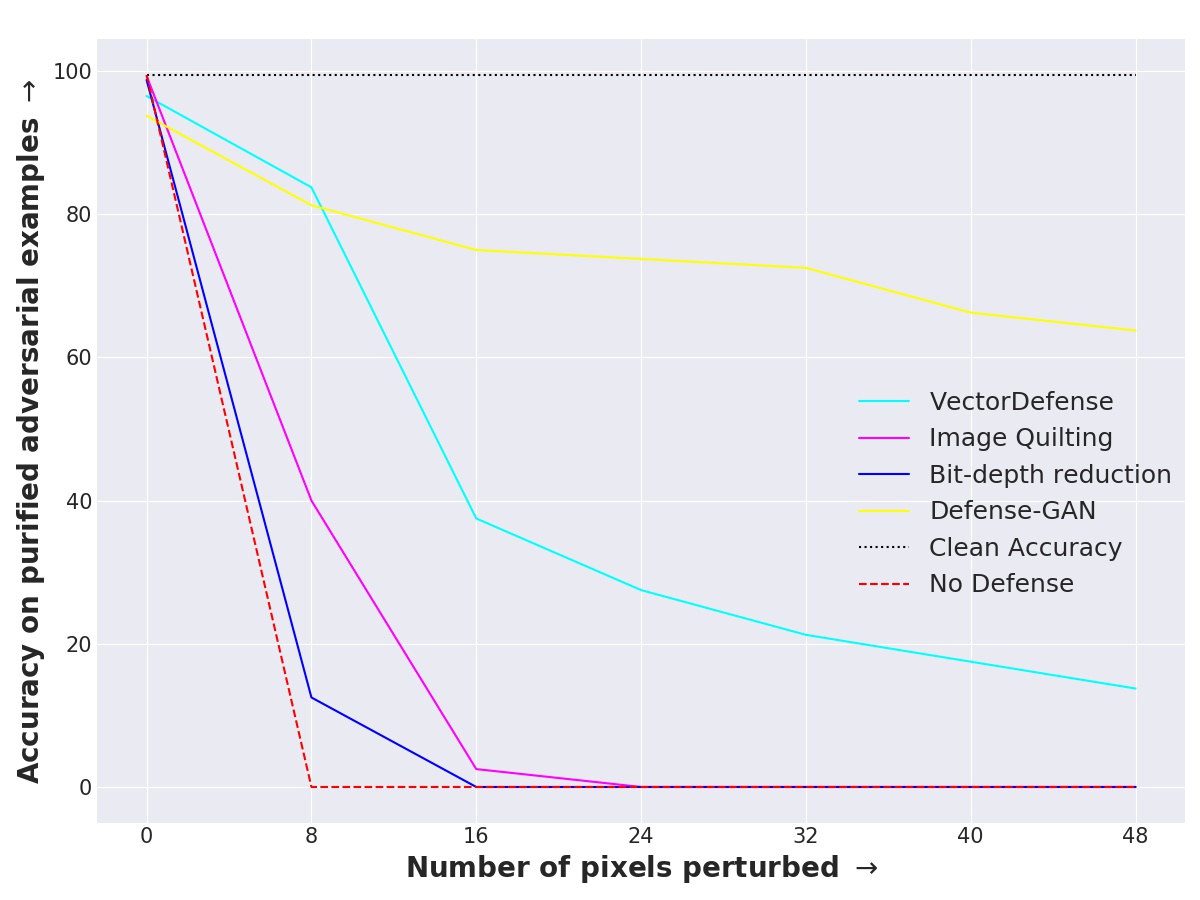}}\label{fig:ba_jsma}}%
	\qquad
	\subfloat[Budget-aware C\&W $L_0$ Attack]{{\includegraphics[width=1.0\columnwidth]{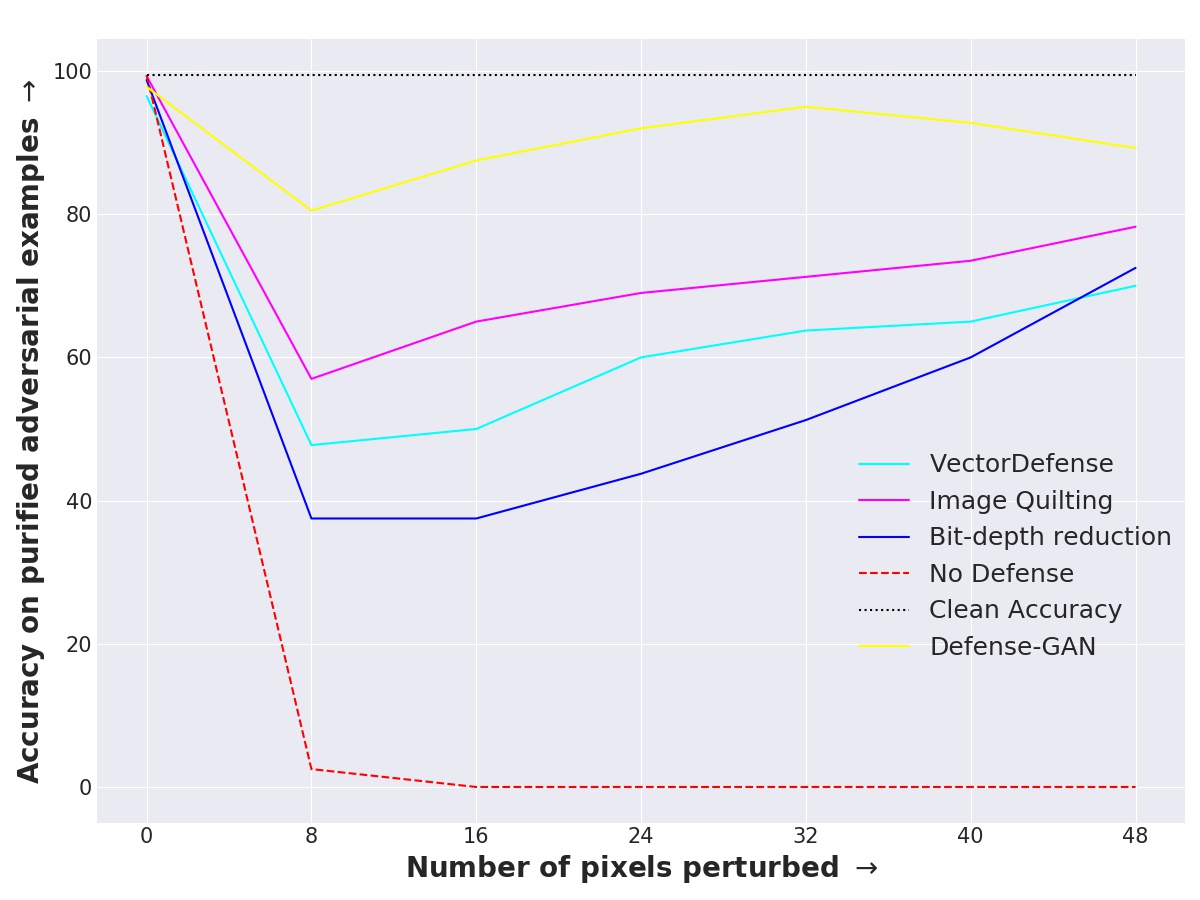}}\label{fig:ba_l0}}
	\caption{Classification accuracy of MNIST classifier (Table \ref{table:mnist_arch}) on purified adversarial examples, crafted using (a) Budget-aware JSMA algorithm and (b) Budget-aware C\&W $L_0$ algorithm and purified by four input transformations (1) VectorDefense; (2) Image quilting \cite{guo2017countering}; (3) Bit-depth reduction \cite{xu2017feature}; and (4) Defense-GAN \cite{samangouei2018defense}. Black dotted lines show the classifier accuracy on clean images, which serves as an upper bound to the effectiveness of a defense. A budget-setting of zero corresponds to the classification accuracy on clean images. Higher is better. }%
	\label{fig:plots}%
\end{figure*}

\noindent\textbf{JSMA Results}
Under budget-aware JSMA, VectorDefense outperformed  bit-depth reduction and image quilting by a large margin ($\ge$17\%; Fig.~\ref{fig:ba_jsma}).
JSMA poses a challenge to input transformation methods by setting the perturbed pixels to the extreme positive (here, solid white; Fig.~\ref{fig:jsma_qual} heatmaps).
The despeckling process is hypothesized to help VectorDefense remove adversarial perturbations more explicitly and effectively than bit-depth reduction and quilting (Fig.~\ref{fig:jsma_qual}).
However, Defense-GAN is still the most effective in recovering from JSMA attacks. 

\noindent\textbf{C\&W~$L_0$ Results} 
Under budget-aware C\&W~$L_0$ attack, VectorDefense performs similarly to the existing hand-designed input transformations across increasing budget settings (Fig.~\ref{fig:ba_l0}).
Note that as the number of pixels perturbed increases, the average perturbation per pixel decreases (Fig.~\ref{fig:cwl0_qual} heatmaps), farther away from the extreme values.
This makes it easier for all hand-designed methods to recover from, leading to (1) the increasing accuracy scores as the budget increases (Fig.~\ref{fig:ba_l0}); and (2)  similar performances across VectorDefense, bit-depth reduction and image quilting (Fig.~\ref{fig:ba_l0}).
Overall, Defense-GAN performed superior compared to hand-designed defenses under general $L_0$ attacks with extreme positive and negative perturbations (Fig.~\ref{fig:cwl0_qual}b--d \& Table~\ref{tbl:gray-results}g).

\subsection{White-Box: Bypassing Input Transformations}\label{sec:exp_white}
Recent research showed that all state-of-the-art defenses are not completely robust to white-box attacks where the attacker has access to the defense mechanism \cite{athalye2018obfuscated}. 
While VectorDefense is not an exception under the white-box Backward Pass Differential Approximation (BPDA) attack \cite{athalye2018obfuscated}, we observed that a large amount of distortion is required to fool the target DNN with VectorDefense.

\noindent\textbf{Experiment} 
Here we employ BPDA~\cite{athalye2018obfuscated} to bypass VectorDefense. 
Basically, BPDA has access to a black-box input transformation method (here, VectorDefense) and uses it to approximate the gradient to compute AXs.
We compare results of running BDPA to attack the target DNN with no defense vs. with VectorDefense under the same experimental setup.
Optimization runs for $50$ steps. 

$L_2$ dissimilarity often describes the average distortion necessary to fool a classifier \cite{athalye2018obfuscated}.
Here, we report the average
$L_2$ dissimilarity for the successfully crafted AXs. 




\noindent\textbf{Results} Though the adversary ends up being successful, VectorDefense makes the BPDA distort the images much more vs. when there is no defense ($L_2$ distortion of $5.30$ vs. $2.97$). Fig. \ref{fig:bpda_qual} shows a qualitative comparison.




\subsection{Hand-designed vs. Learned Prior}
\label{sec:exp_inter}


Throughout the previous results, we have qualitatively seen the superiority of Defense-GAN---an input transformation method with a learned prior---compared to hand-designed methods including VectorDefense.
We further compare and contrast the process of purifying AXs by these two types of input transformation approaches.

Under strong attacks with 40-pixel perturbations, Defense-GAN substantially outperformed all hand-designed transformations 
(Fig.~\ref{fig:plots}; yellow lines at 40).

To shed more light into this result, we qualitatively examine cases when Defense-GAN succeeded but others failed and showed interesting observations in Fig.~\ref{fig:interpretable_qual}.
Here, the images are strongly perturbed such that the AXs actually qualitatively change into the target class (Fig.~\ref{fig:interpretable_qual} row 3; digit ``6'' after perturbation does look like a ``0''). 
However, Defense-GAN prior is so strong that it pulls the image back into the correct class ``6'' (Fig.~\ref{fig:interpretable_qual}d). 
In contrast, VectorDefense removes the speckles around the digit and turns the image into a ``0''---the label that may make sense to humans but is not the original label.
Similar opposite behaviors of Defense-GAN vs. VectorDefense can be observed in other cases (Fig.~\ref{fig:interpretable_qual}).

\section{Discussion and Future Work}\label{sec:conclusion}

We present VectorDefense---a novel defense to AXs that transforms a
bitmap input image into the space of vector graphics and back, prior to
classification. 
We showed how this transformation smooths out the visual artifacts created by state-of-the-art attack algorithms.


Under strong perturbations, VectorDefense and other hand-designed input transformation methods underperformed Defense-GAN, a method with a strong, learned prior.
While this result may not be too surprising, 
we view VectorDefense as a stepping stone towards decomposing images into compact, interpretable elements to solve the AX problem.
Simple use of vectorization in this work alone reduces effectiveness of AXs, and opens up promising extensions towards more robust machine learning:

\begin{itemize}
	\item The vector representation (SVG) could be further compacted via Ramer-Peucker-Douglas algorithm \cite{douglas1973algorithms, ramer1972iterative} or constrained to compose of elements from a strict set of geometric primitives (e.g. only straight strokes).
	\item Since the vector images are resolution-independent, one could rasterize them back into much smaller-sized images. In lower-dimensional space, it can be easier to defend against AXs \cite{gilmer2018adversarial}.
	\item Another extension of VectorDefense would be to learn a generative model on this vector space.
	\item In high-dimensional color image space, simple image tracing may not yield compact and interpretable elements (Sec.~\ref{sec:color}); therefore, we did not explore further vectorizing natural images. 
	Instead it might be interesting to de-render an image into a scene graph \cite{wu2017neural}, and train a prior over the graphs \cite{krishnavisualgenome}.
\end{itemize}



\subsection*{Acknowledgements}
We thank Zhitao Gong, Chengfei Wang for feedback on the drafts; and Nicholas Carlini and Nicolas Papernot for helpful discussions.

\newpage
{\small
\bibliographystyle{ieee}
\bibliography{egbib}
}

\clearpage

\renewcommand{\thesection}{S\arabic{section}}
\renewcommand{\thesubsection}{\thesection.\arabic{subsection}}

\newcommand{\beginsupplementary}{%
	\renewcommand{\thetable}{S\arabic{table}}%
	\renewcommand{\thefigure}{S\arabic{figure}}%
}
\newcommand{\suptitle}{\hspace{4.2em} Supplementary materials for: \newline VectorDefense: Vectorization as a Defense to Adversarial Examples}

\newcommand{\toptitlebar}{
	\hrule height 4pt
	\vskip 0.25in
	\vskip -\parskip%
}
\newcommand{\bottomtitlebar}{
	\vskip 0.29in
	\vskip -\parskip
	\hrule height 1pt
	\vskip 0.09in%
}

\beginsupplementary

\newcommand{\maketitlesupp}{
	\newpage
	\twocolumn[
	\begin{@twocolumnfalse}
		\null
		\vskip .375in
		\begin{center}
			{\Large \bf \suptitle \par}
			\vspace*{24pt}
			{
				\large
				\lineskip .5em
				\par
			}
			\vskip .5em
			\vspace*{12pt}
		\end{center}
	\end{@twocolumnfalse}
	]
}

\maketitlesupp

\section{Tracing color images}
\label{sec:color}
\begin{figure}[h]
	\begin{center}
		\hspace{0.0em}
		SVHN
		\hspace{8.0em}
		CIFAR-10				
		\includegraphics[width=0.48\textwidth]{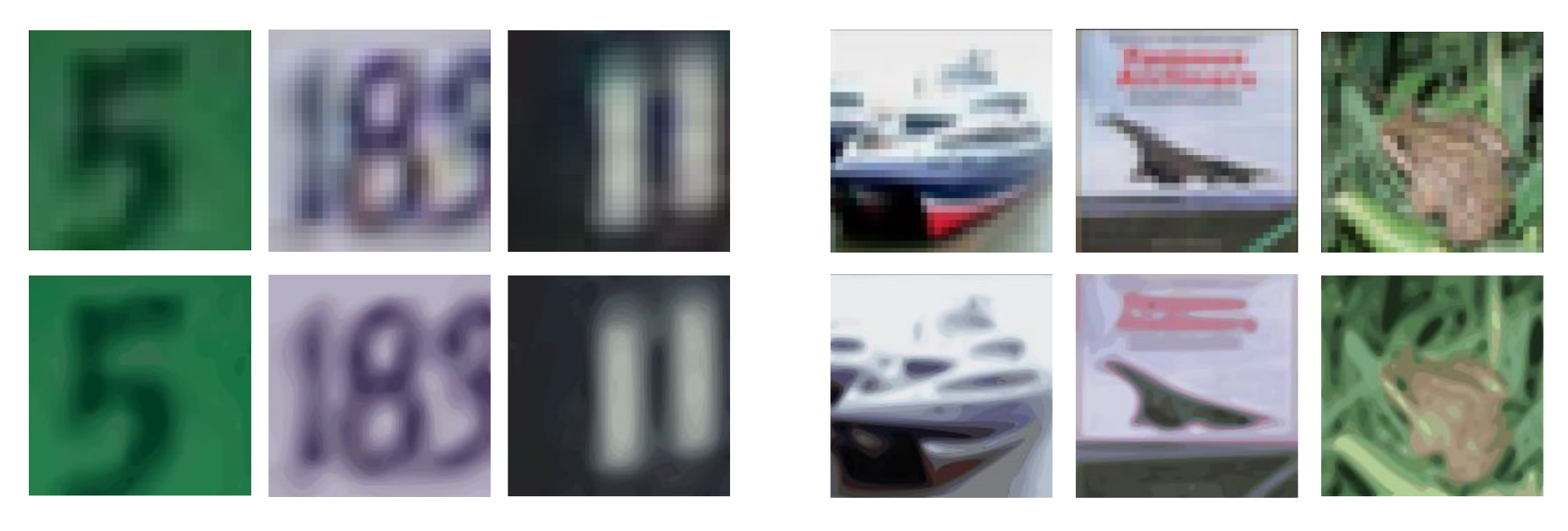}
		\caption{Tracing color images using Potrace: Top row shows test images in the bitmap space. Bottom row shows the corresponding vector graphic outputs.
		}
		\label{fig:colortrace_image}
	\end{center}
	
\end{figure}

Fig.~\ref{fig:colortrace_image} shows samples from SVHN and CIFAR-10 datasets traced via Potrace (each RGB channel being traced independently). 
While color images can be traced via vectorization (Fig.~\ref{fig:colortrace_image}), the output vector are often not composed of simple compact, interpretable geometric primitives. Therefore, we did not explore it further in this work.

\section{Experimental Setup}
\label{sec:experimental_setup}
\noindent\textbf{VectorDefense}
To vectorize the
input image, we used the open-source implementation of the Potrace
algorithm\footnote{\url{http://potrace.sourceforge.net/}}. In our experiments, we make two changes to the default potrace algorithm.
\begin{itemize}
	\item We set the \emph{turn policy} \cite{selinger2003potrace} to black, which prefers to connect the black components in Step 1 (Sec. \ref{sec:potrace}).
	\item We increased the default maximum size of
	removed speckles, the \textit{despeckling} value, to $5$ (Sec. \ref{sec:vectordefense}).
\end{itemize}
With the image converted to a scalable vector graphics
(SVG) format, we render it with the open-source
Inkscape\footnote{\url{https://inkscape.org/}} program, having the same width
and height as the original image. 

\noindent\textbf{Bit-depth reduction}
We reduce the images to 1-bit in our experiments.

\noindent\textbf{Image-quilting}
In our implementation of image quilting on MNIST, we use a quilting patch size of 4x4 and a database of 60,000 images from the MNIST training set. The patch is selected at random from one of $K=10$ nearest neighbours. 

\noindent\textbf{Defense-GAN} 
We set the parameters of L and R to $200$ and $10$ respectively.

\begin{table}[b]
	\caption{\textbf{Neural network architecture used for 
			the MNIST dataset.} 
		Conv: convolutional layer, FC: fully connected layer.\\}
	\centering
	\begin{tabular}{@{} c c c c@{}c @{}}
		\toprule
		\textbf{CONFIGURATION}\\
		\midrule
		Conv(64, 5, 5) + Relu \\
		Conv(64, 5, 5) + Relu \\
		Dropout(0.25) \\
		FC(128) + Relu \\
		Dropout(0.5) \\
		FC + Softmax \\
		\midrule
	\end{tabular}
	\label{table:mnist_arch}
\end{table}

\section{Additional results by VectorDefense}
\label{sec:app_ex}

\begin{figure*}[th]
	\vspace*{-0.8em}
	\begin{center}
		\includegraphics[width=2.0\columnwidth]{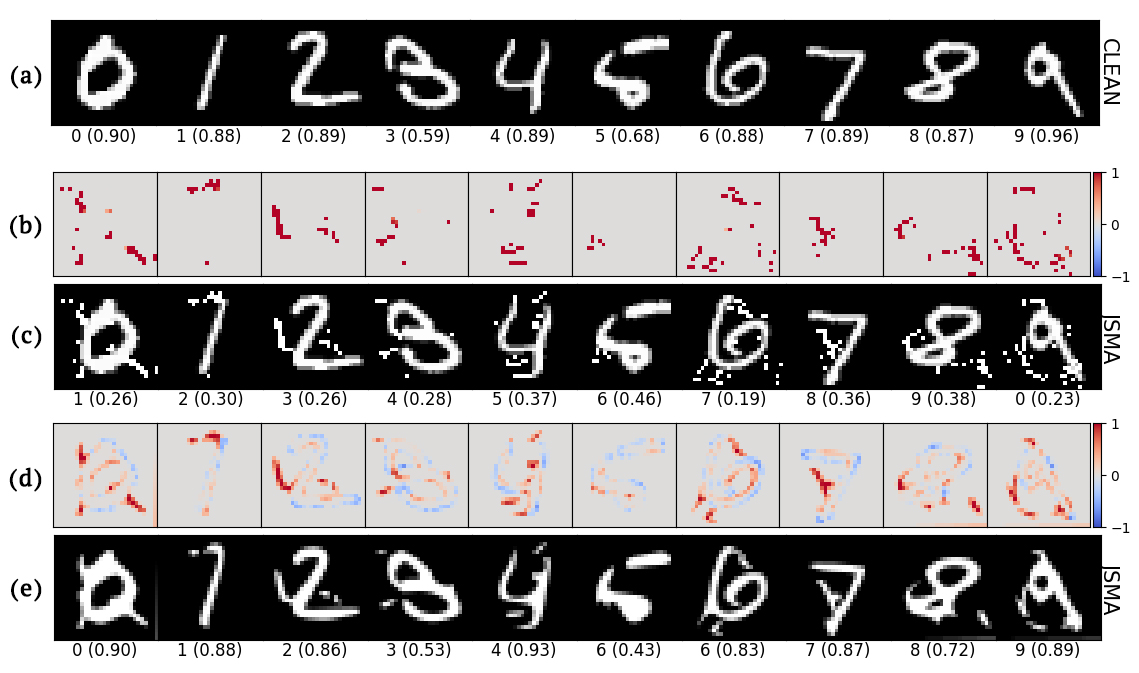}
		\caption{JSMA: Given starting images (a), adversarial images are generated using JSMA (c). VectorDefense effectively purifies the adversarial images (e). The perturbations introduced by the adversary (b) and the purification process (d) can be obtained by subtracting (a) from (c) and (a) from (e) respectively. All images start out classified correctly with label $l$ and are targeted to have a label $l+1$. The images were chosen as the first of their class from the MNIST test set.
		}
		\label{fig:hm_jsma}
	\end{center}
\end{figure*}

\begin{figure*}[th]
	\vspace*{-0.8em}
	\begin{center}
		\includegraphics[width=2.0\columnwidth]{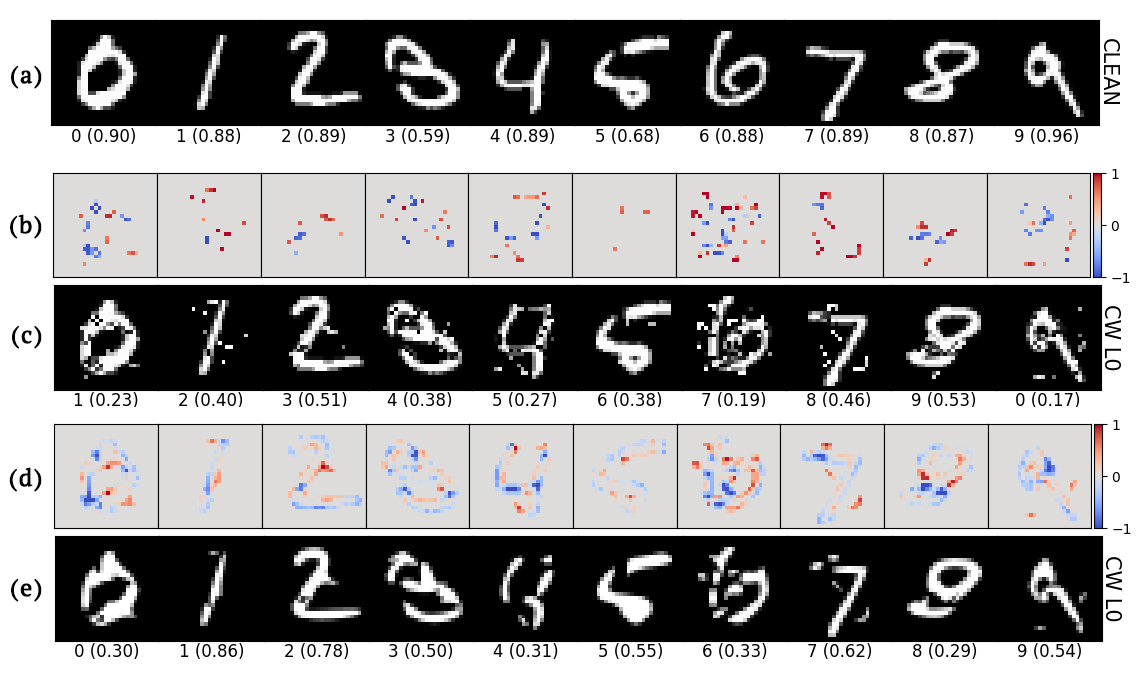}
		\caption{C\&W $L_0$: Given starting images (a), adversarial images are generated using C\&W $L_0$ (c). VectorDefense effectively purifies the adversarial images (e). The perturbations introduced by the adversary (b) and the purification process (d) can be obtained by subtracting (a) from (c) and (a) from (e) respectively. All images start out classified correctly with label $l$ and are targeted to have a label $l+1$. The images were chosen as the first of their class from the MNIST test set.
		}
		\label{fig:hm_cwl0}
	\end{center}
\end{figure*}

\begin{figure*}[th]
	\vspace*{-0.8em}
	\begin{center}
		\includegraphics[width=2.0\columnwidth]{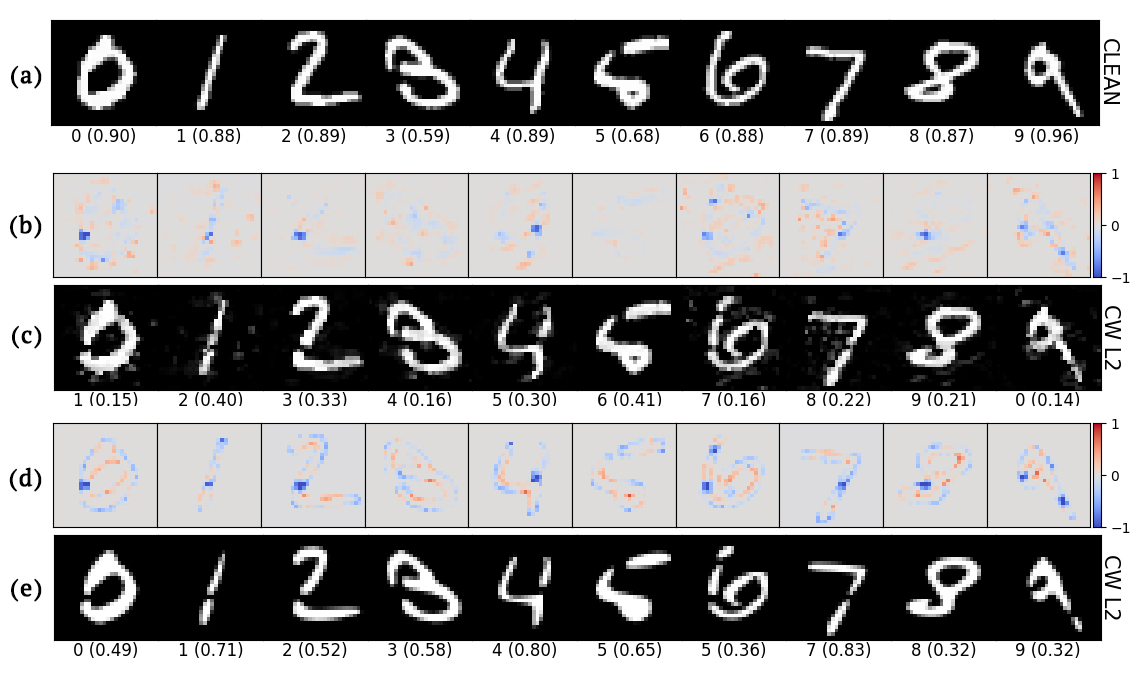}
		\caption{C\&W $L_2$: Given starting images (a), adversarial images are generated using C\&W $L_2$ (c). VectorDefense effectively purifies the adversarial images (e). The perturbations introduced by the adversary (b) and the purification process (d) can be obtained by subtracting (a) from (c) and (a) from (e) respectively. All images start out classified correctly with label $l$ and are targeted to have a label $l+1$. The images were chosen as the first of their class from the MNIST test set.
		}
		\label{fig:hm_cwl2}
	\end{center}
\end{figure*}

\begin{figure*}[th]
	\vspace*{-0.8em}
	\begin{center}
		\includegraphics[width=2.0\columnwidth]{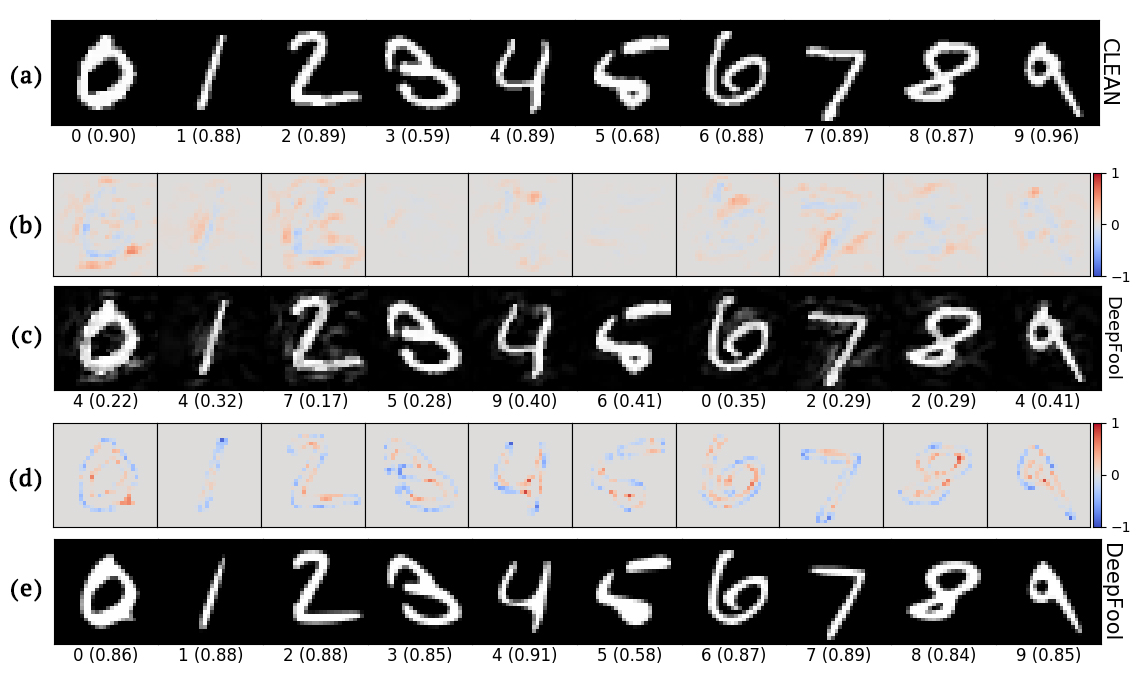}
		\caption{DeepFool: Given starting images (a), adversarial images are generated using DeepFool (c). VectorDefense effectively purifies the adversarial images (e). The perturbations introduced by the adversary (b) and the purification process (d) can be obtained by subtracting (a) from (c) and (a) from (e) respectively. All images start out classified correctly with label $l$ and are targeted to have a label $l+1$. The images were chosen as the first of their class from the MNIST test set.
		}
		\label{fig:hm_deepfool}
	\end{center}
\end{figure*}

We show below five figures demonstrating how VectorDefense purifies AXs created by five different attack algorithms: I-FGSM (Fig.~\ref{fig:hm_ifgsm}), JSMA (Fig.~\ref{fig:hm_jsma}), C\&W $L_0$ (Fig.~\ref{fig:hm_cwl0}), C\&W $L_2$ (Fig.~\ref{fig:hm_cwl2}), and DeepFool (Fig.~\ref{fig:hm_deepfool}).

%
%

\begin{figure*}[th]
	\begin{center}
		\includegraphics[width=2.0\columnwidth]{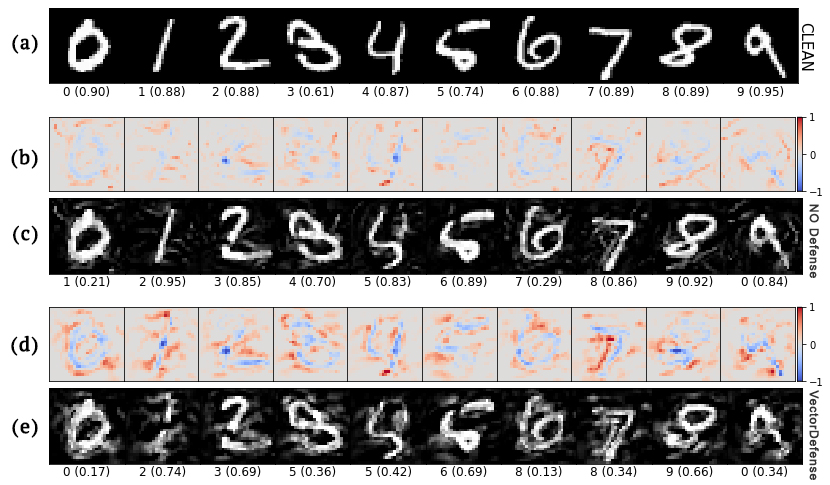}
		\caption{Distortion introduced by BPDA against VectorDefense. Given starting images (a), adversarial images are generated using PGD (c) corresponding to a $L_2$ dissimilarity of 2.97. Once VectorDefense is augmented to the image classification pipeline, the adversarial images are then generated using PGD with BPDA (e) corresponding to a $L_2$ dissimilarity of 5.30. The perturbations introduced by the adversary under no defense (b) and the adversary in the presence of VectorDefense (d) can be obtained by subtracting (a) from (c) and (a) from (e) respectively. All images start out classified correctly with label $l$ and are targeted to have a label $l+1$. The images were chosen as the first of their class from the MNIST test set.
	 }
		\label{fig:bpda_qual}
	\end{center}
\end{figure*}

\begin{figure*}[t]
	\begin{center}
		Number of pixels perturbed in our $L_0$ attack
	\end{center}
	\hspace*{13.5em}
	(a) \textbf{8}
	\hspace{1.0em}
	(b) \textbf{16}
	\hspace{1.0em}
	(c) \textbf{24}
	\hspace{0.8em}
	(d) \textbf{32}
	\hspace{0.5em}
	(e) \textbf{40}
	\hspace{0.5em}
	(g) \textbf{48}
	\begin{center}
		\includegraphics[width=0.48\textwidth]{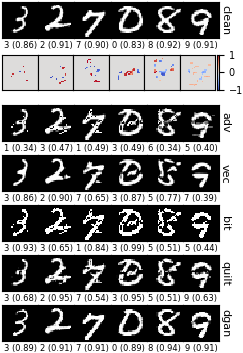}
		\caption{Qualitative comparison of input transformations on budget-aware C\&W $L_0$. This figure follows the same conventions as in Fig. \ref{fig:jsma_qual}. 
		Row 2: the intensities of perturbations gradually decrease as we move from the least (a) to the greatest number of pixels (g). Hand-designed input transformation methods can only remove, but cannot fill in missing pixels as Defense-GAN (bottom row). 
		}
		\label{fig:cwl0_qual}
	\end{center}
\end{figure*}

\end{document}